\documentclass[sigconf]{acmart}

\AtBeginDocument{%
  \providecommand\BibTeX{{%
    \normalfont B\kern-0.5em{\scshape i\kern-0.25em b}\kern-0.8em\TeX}}}

\copyrightyear{2021}
\acmYear{2021}
\setcopyright{acmcopyright}\acmConference[MM '21]{Proceedings of the 29th ACM International Conference on Multimedia}{October 20--24, 2021}{Virtual Event, China}
\acmBooktitle{Proceedings of the 29th ACM International Conference on Multimedia (MM '21), October 20--24, 2021, Virtual Event, China}
\acmPrice{15.00}
\acmDOI{10.1145/3474085.3475271}
\acmISBN{978-1-4503-8651-7/21/10}

\usepackage{kotex}
\usepackage{subcaption}
\usepackage{multirow}
\usepackage{ulem}

\definecolor{darkolivegreen}{rgb}{0.33, 0.42, 0.18}
\definecolor{darkpastelgreen}{rgb}{0.01, 0.75, 0.24}
\definecolor{mediumjunglegreen}{rgb}{0.11, 0.21, 0.18}

\newcommand{\ie}{\textit{i}.\textit{e}.}

\newcommand{\etal}{\textit{et} \textit{al}.}

\newcommand{\Fref}[1]{Fig.~\ref{#1}}

\newcommand{\Tref}[1]{Table~\ref{#1}}

\acmSubmissionID{567}

\begin{document}

\title{Feature Stylization and Domain-aware Contrastive Learning\\for Domain Generalization}

\renewcommand{\shorttitle}{Feature Stylization and Domain-aware Contrastive Learning for Domain Generalization}


\author{Seogkyu Jeon}
\affiliation{
    \institution{Department of Computer Science}
    \city{Yonsei University}
    \country{South Korea}
}
\email{jone9312@yonsei.ac.kr}

\author{Kibeom Hong}
\affiliation{
    \institution{Department of Computer Science}
    \city{Yonsei University}
    \country{South Korea}
}
\email{cha2068@yonsei.ac.kr}

\author{Pilhyeon Lee}
\affiliation{
    \institution{Department of Computer Science}
    \city{Yonsei University}
    \country{South Korea}
}
\email{lph1114@yonsei.ac.kr}

\author{Jewook Lee}
\affiliation{
    \institution{Department of Computer Science}
    \city{Yonsei University}
    \country{South Korea}
}
\email{hooraid@yonsei.ac.kr}

\author{Hyeran Byun}
\authornote{Corresponding Author}
\authornote{Also with Graduate school of Artificial Intelligence, Yonsei University}
\affiliation{
    \institution{Department of Computer Science}
    \city{Yonsei University}
    \country{South Korea}
}
\email{hrbyun@yonsei.ac.kr}



\begin{abstract}
Domain generalization aims to enhance the model robustness against domain shift without accessing the target domain. Since the available source domains for training are limited, recent approaches focus on generating samples of novel domains. Nevertheless, they either struggle with the optimization problem when synthesizing abundant domains or cause the distortion of class semantics. To these ends, we propose a novel domain generalization framework where feature statistics are utilized for stylizing original features to ones with novel domain properties. To preserve class information during stylization, we first decompose features into high and low frequency components. Afterward, we stylize the low frequency components with the novel domain styles sampled from the manipulated statistics, while preserving the shape cues in high frequency ones. As the final step, we re-merge both the components to synthesize novel domain features.
To enhance domain robustness, we utilize the stylized features to maintain the model consistency in terms of features as well as outputs.
We achieve the feature consistency with the proposed domain-aware supervised contrastive loss, which ensures domain invariance while increasing class discriminability.
Experimental results demonstrate the effectiveness of the proposed feature stylization and the domain-aware contrastive loss.
Through quantitative comparisons, we verify the lead of our method upon existing state-of-the-art methods on two benchmarks, PACS and Office-Home.
\end{abstract}




\maketitle

\section{Introduction}
Since the remarkable advance of deep neural networks, they have become ubiquitous in various fields, especially computer vision systems. However, there still exist potential risks lying on the flip side of their success. One of the main concerns is their vulnerability against visual domain shift~\cite{ben2007analysis, ben2010theory, saenko2010adapting}. Concretely, deep models react unexpectedly when confronting data unaffiliated to the training distribution. For example, an auto-tagging model trained on clean product images shows poor performance when taking as inputs real product images which are under various viewpoints and light conditions.

To equip the models with the ability in coping with domain shift, previous studies tackle the problem of domain adaptation~\cite{daume2009frustratingly, sun2016return, tzeng2017adversarial, saito2018maximum, yue2019domain, kim2020learning}.
In this problem setting, two different domains sharing the same label space are prepared for training and test, which are referred to as the ``source'' and the ``target'' domains, respectively.
During training, a model has access to both labeled images from the source domain and unlabeled (or partially labeled) images from the target domain.
Being aware of the target domain, existing works successfully minimize the discrepancy between two distinct domains, thus leading to large performance boosts~\cite{ben2010theory, tzeng2017adversarial, sun2016return, tzeng2017adversarial, hoffman2018cycada, saito2018maximum, yue2019domain, kim2020learning, Xu2018DeepCN}.

However, the problem setting of domain adaptation is impractical in that domain shift is generally unpredictable in the real-world scenarios, \ie, we do not know the target domain at training time.
To this end, a new task has attracted much attention recently, aiming to learn domain robustness without accessing the target domain data, namely domain generalization~\cite{carlucci2019domain, Li2018LearningTG, Li2019EpisodicTF, dou2019domain, li2018domain, ghifary2015domain, Seo2020LearningTO, Chattopadhyay2020LearningTB, huang2021fsdr, qiao2021uncertainty, choi2021robustnet, xu2021fourier}.
In this setting, multiple datasets from different source domains are typically utilized to learn domain invariant representations. 

Previous methods~\cite{zhou2020learning, nuriel2020permuted, xu2020robust, qiao2020learning} embrace the observation that domain robustness is proportional to the number of domains observable in the training stage~\cite{tobin2017domain}.
To that end, they utilize generative adversarial networks (GAN)~\cite{zhou2020learning} or adaptive instance normalization~(AdaIN)~\cite{huang2017arbitrary, nuriel2020permuted} for synthesizing novel (unseen) domains.
Nonetheless, they have two clear limitations which are critical for domain generalization.
First, GAN-based methods become prohibitively difficult to optimize as the number of novel domains increases, which limits the size of observable domain space.
Next, AdaIN-based approaches fail to preserve the semantics of original images, as instance normalization~(IN) tends to wash away class discriminative information~\cite{nam2018batch, Seo2020LearningTO}.

In this paper, we introduce a novel framework for domain generalization, overcoming the above limitations.
Specifically, to synthesize novel domains without losing class discriminative information, we propose a novel \textit{feature stylization block}.
First, we calculate batch-wise feature statistics of source domains and sample novel domain styles from the feature distribution.
We re-scale the standard deviation of the source feature distribution so that the outlying style statistics are more likely to be sampled.
However, the original semantics can be distorted during the stylization process, which will disturb the training. 
To preserve the original semantics during stylization, inspired by a recent photo-realistic stylization method~\cite{Yoo2019PhotorealisticST}, we decompose original features into high and low frequency components which contain structural and textural information, respectively.
Afterwards, we manipulate the low frequency components while remaining shape cues in high frequency ones to prevent semantics distortion.
Lastly, we re-merge the stylized low frequency components and the high frequency ones, leading to the stylized features.
By incorporating them in the training, our model is allowed to learn robust representation against domain shift.

Rather than naively utilizing stylized features for training, we seek for better strategies that can provide domain robustness guidance.
Intuitively, a robust model against domain shift should yield consistent predictions for the stylized features and the original ones. In this point of view, we adopt the \textit{consistency loss} to maximize the agreement between the model predictions for them.
Concretely, we measure the KL divergence between two output distributions and minimize it with the consistency loss.

Moreover, we propose the \textit{domain-aware supervised contrastive loss} to minimize distance between the stylized and the original features, in order to achieve feature-level consistency. Although the conventional supervised contrastive loss has proven to be effective, we found that it is unsuitable for domain generalization.
The loss expels the samples from different domains and thus disturbs domain invariance, which conflicts with the goal of domain generalization.
To this end, we introduce the novel domain-aware supervised contrastive loss which ignores negative samples from different domains, hence preserving domain invariance while empowering class discriminability.

The contributions of this paper can be summarized into three folds. Firstly, we propose a novel domain generalization framework, where diverse domain styles are generated and leveraged through the proposed feature stylization block.
The stylized features are in turn used to enhance domain robustness by encouraging the model to produce consistent outputs.
Secondly, we introduce the novel domain-aware supervised contrastive loss. The proposed loss strengthens the domain invariance by contrasting features with respect to domain and class labels. Lastly, we demonstrate the effectiveness of each component of our model through analyses and ablation studies. Furthermore, experimental results show that our method surpasses previous methods with obvious margins, achieving a new state-of-the-art on the widely used benchmarks: PACS and Office-Home. Even on the single-source domain generalization task, our method shows delightful performance improvements over the baseline.

\section{Related Works}

\subsection{Domain Adaptation}
Domain adaptation aims to transfer learned knowledge from source domains to a target domain. In this setting, the source domain is usually a large scale dataset with annotations, and the target domain data is either partially labeled or completely unlabeled. They are referred to semi-supervised domain adaptation~(SSDA)~\cite{donahue2013semi, yao2015semi, ao2017fast, saito2019semi} and unsupervised domain adaptation~(UDA)~\cite{daume2009frustratingly, sun2016return, tzeng2017adversarial, hoffman2018cycada, saito2018maximum, yue2019domain, kim2020learning}, respectively. 

Semi-supervised domain adaptation methods impose constraints on both labeled and unlabeled instances of the target domain in various ways. Donahue~\etal~\cite{donahue2013semi} build a similarity graph to constrain unlabeled data and transfer knowledge with a projective model transfer method. Ao~\etal~\cite{ao2017fast} distill knowledge from the source domain by generating pseudo labels for the unlabeled target data. Saito~\etal~\cite{saito2019semi} estimate class-specific prototypes with sparsely labeled examples of the target domain, then update them by solving a minimax game on the unlabeled data.

In unsupervised domain adaptation, most methods~\cite{daume2009frustratingly, sun2016return, tzeng2017adversarial, hoffman2018cycada, kim2020learning} conduct feature alignment between source and target domains. To this end, CORAL~\cite{sun2016return} minimizes the distance between the covariance matrices, while ADDA~\cite{tzeng2017adversarial} employs a domain discriminator for adversarial learning. Meanwhile, CyCADA~\cite{hoffman2018cycada} adopts an image-to-image translation framework to transfer the source domain data to the target domain data on image-level.
Recently, domain randomization~\cite{tobin2017domain, yue2019domain, zakharov2019deceptionnet, kim2020learning} is another generative stream which diversifies the textures of source domain images, allowing the model to learn texture invariant representations. Yue~\etal~\cite{yue2019domain} manipulate images into an external class from ImageNet~\cite{krizhevsky2012imagenet}, and LTIR~\cite{kim2020learning} exploits an artistic style transfer method to alter the textures of the source and target domains.

Our method relates to the domain randomization approaches in that it aims to generate features with diverse domain characteristics.
However, they are not suitable for domain generalization as they require external datasets~\cite{krizhevsky2012imagenet, yue2019domain, kim2020learning}.
On the contrary, our proposed feature stylization block is able to generate various stylized features based on the statistics of source domains, without access to additional data.

\begin{figure*}
  \includegraphics[width=0.92\textwidth]{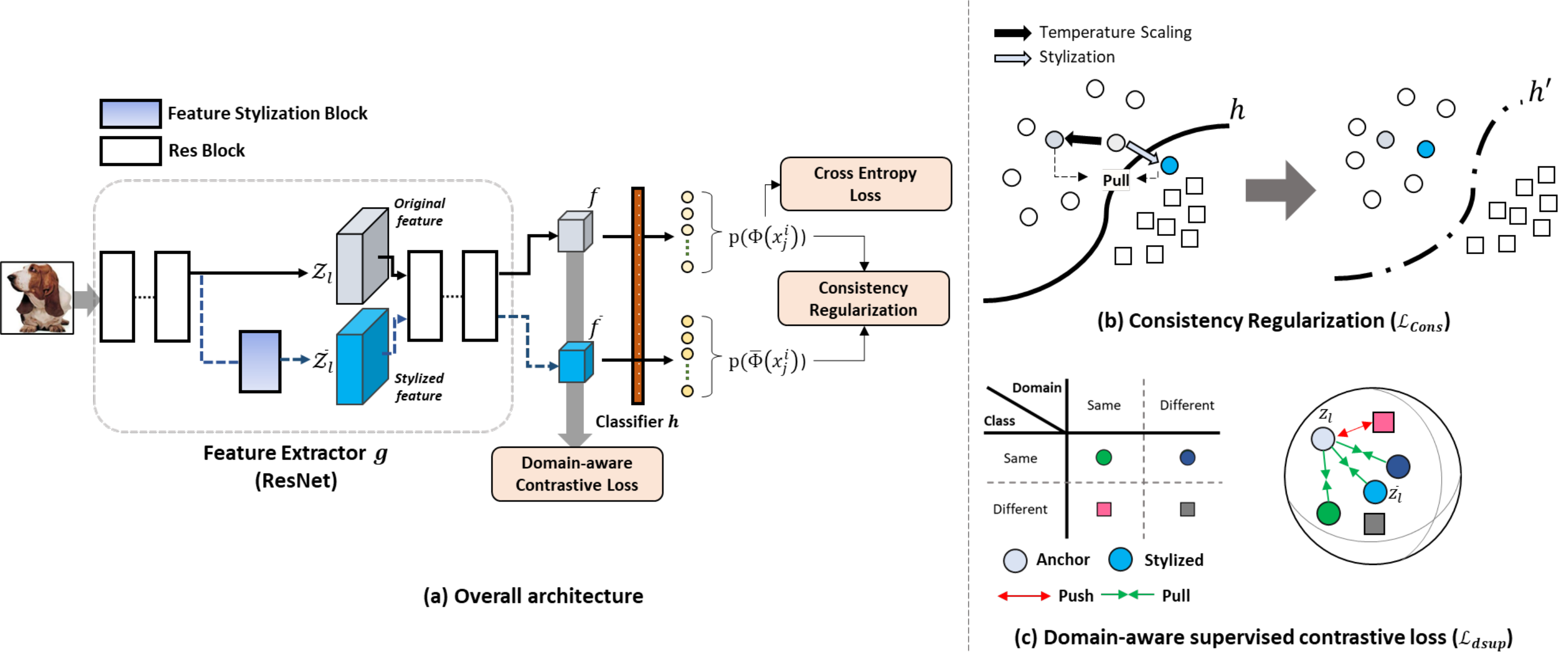}
  \caption{The overview of our proposed methods. We illustrate the overall architecture on (a). The backbone consists of multiple convolutional layers, and we stylize the feature on the intermediate layer of backbone. Both origianl and stylized features follow the same forward path, producing class predictions $p(\Phi(x^i_j))$ and $p(\Bar{\Phi}(X^i_j))$. The predictions and features are exploited with the consistency regularization~($\mathcal{L}_{cons}$) and domain-aware supervised contrastive loss~($\mathcal{L}_{dsup}$) which are described in (b) and (c), respectively.}
  \label{fig:Arch}
\end{figure*}

\subsection{Domain Generalization}

The goal of domain generalization is to learn domain invariant representations based on only source domains. Different from unsupervised domain adaptation, target domain data is inaccessible during training, making the task more challenging.
In addition, multiple domains are typically utilized to achieve domain-agnostic representation without having target domain data.
Previous domain generalization methods can be roughly categorized into four groups: meta-learning, architectural modification, regularization, and generative approaches.

The first group exploits meta-learning techniques~\cite{finn2017model, ravi2016optimization} to align different domains~\cite{Li2018LearningTG, balaji2018metareg, Li2019EpisodicTF, dou2019domain}. These approaches borrow the powerful adaptability of meta-learning algorithms whose effectiveness is proven in the field of few-shot learning. Representatively, Li~\etal~\cite{Li2019EpisodicTF} separates the training set into multiple episodes, each of which handles only a single domain.
During training, they update the backbone with aggregated regularization losses from domain specific networks.
Meanwhile, MASF~\cite{dou2019domain} simulates domain shift using different episodes. They perform global alignment of class relationships while clustering local samples.

Secondly, some works~\cite{Motiian2017UnifiedDS, li2018domain, ghifary2015domain, Seo2020LearningTO, Chattopadhyay2020LearningTB, xu2021fourier} try architectural changes to model a shared embedding space~\cite{Motiian2017UnifiedDS, li2018domain, ghifary2015domain} or to build domain-specific networks~\cite{Chattopadhyay2020LearningTB, Seo2020LearningTO}. Exploiting auxiliary pretext tasks are also favored as a sub-stream~\cite{carlucci2019domain, Wang2020LearningFE}. As a pioneer, JiGen~\cite{carlucci2019domain} proposes to solve jigsaw puzzles as an auxiliary task to induce the model to learn the concepts of spatial correlation. Inheriting from JiGen, EIS-Net~\cite{Wang2020LearningFE} employs a momentum metric learning task to provide extrinsic relationship supervision. 
Other approaches~\cite{wang2018learning, wang2019learning, Huang2020SelfChallengingIC, shi2020informative, choi2021robustnet} apply diverse regularization during training. HEX~\cite{wang2018learning} employs the neural gray-Level co-occurrence matrix to find superficial representations related to the task. PAR~\cite{wang2019learning} penalizes the predictive power of earlier layers so that the model relies more on global representations from deeper layers. RSC~\cite{Huang2020SelfChallengingIC} masks out both spatial regions and channels which have high contributions to the task. RobustNet~\cite{choi2021robustnet} encourage model to utilize domain-invariant features by selectively whitening domain-variant feature channels in the gram matrix during training.

Lastly, based on the intuition that the generalization ability can be boosted with samples from more diverse domains~\cite{tobin2017domain}, generative approaches arises~\cite{zhou2020learning, nuriel2020permuted, xu2020robust, qiao2020learning}. They augment the training set with samples similar in semantics but different in domain characteristics. L2A-OT~\cite{zhou2020learning} adopts generative adversarial networks to synthesize images which are distant from original ones in terms of the Wasserstein distance. Qiao \etal~\cite{qiao2020learning} apply adversarial perturbations on the images to augment the source domains. From the perspective of frequency, FACT~\cite{xu2021fourier} analyze the frequency components of the image with the fourier transformation, and conduct data augmentation by mixing the amplitude information.

Our method can be viewed as a harmonious combination of the generative method and the regularization-based approach. We generate features of novel domains via the novel feature stylization block during training, then apply regularization in terms of output consistency and feature similarity. The efficacy of our method is demonstrated through extensive experiments in Sec.~\ref{sec:Experiments}.

\section{Methods}

In this section, we first describe the baseline setup of multi-source domain generalization for image classification,  then introduce our novel feature stylization method and consistency learning process thereafter.
The overall framework of our method is illustrated in \Fref{fig:Arch}~(a).

\subsection{Baseline}
In the multi-source domain generalizaiton task, multiple datasets of $K$ source domains $\mathcal{D} = \left \{ D_1, D_2, ... , D_K \right \}$ are accessible during training. Each dataset $D_i$ contains a set of images $ X^i = \left \{ x^i_1, x^i_2, ..., x^i_{n_i}\right \}$ with the corresponding class label set $Y^i = \left \{ y^i_1, y^i_2, ..., y^i_{n_i}\right \}$, where $n_i$ is the number of images in the $i$-th dataset. Naturally, the domain label of $x^i_j$ can be obtained as $d_j^i = i$. We also note that all datasets share the same label space, \ie, $y^i_j \in \mathcal{Y}$.
We train a neural network $\Phi$ which consists of a feature extractor $g$ and a following classifier $h$. The feature extractor is composed of multiple convolutional layers and we denote the output features of the $l$-th convolutional layer by $z_{l} \in \mathbb{R}^{B \times C_l \times H_l \times W_l}$, where $B$ is the cardinality of a mini-batch, and $C_l$ is the number of channels, while $H_l$ and $W_l$ are the height and width of the feature, respectively.
The classifier is a single fully-connected layer.
We train the network $\Phi$ by minimizing the cross-entropy loss as follows.
\begin{equation}
    \mathcal{L}_{ce} = - \frac{1}{K}\sum_{i=1}^{K} \frac{1}{n_i} \sum_{j=1}^{n_i}y^i_j\log(p(\Phi(x^i_j))),
\label{equ:loss_cross_entropy}
\end{equation}
where $p(\cdot)$ indicates the softmax function.
Consequently, with the above baseline setup, the network $\Phi$ is trained to classify an image $x^i_j$ into its corresponding label $y^i_j$.

\subsection{Feature Stylization}

An intuitive way to improve the generalization ability of a model would be allowing it to see diverse samples from different domains~\cite{tobin2017domain}. In this point of view, we augment the source domains by synthesizing novel domains by manipulating feature statistics. Before stylization, we note that it should be ensured that the generated feature should maintain the original semantics. To this end, we borrow the feature decomposition of a photo-realistic style transfer model~\cite{Yoo2019PhotorealisticST}, where structural features and textural features are separated into high frequency and low frequency components, respectively. The feature decomposition process is formulated as:

\begin{equation}
\label{equ:feature_decomposition}    
    \begin{aligned}
        z^L_{l} &= \text{UP}(\text{AvgPool}(z_{l})), \\
        z^H_{l} &= z_{l} - z^L_{l},
    \end{aligned}
\end{equation}
where ``AvgPool'' denotes spatial average pooling operation with the kernel size of 2, and ``UP'' indicates nearest neighbor upsampling operation.
After decomposition, we perform stylization on the low frequency feature $z^L_l$ only to preserve structural information.

Since neither extra datasets nor pre-trained networks are available in our setting, we stylize the feature by utilizing its batch-wise statistics. Firstly, the mean and variance are obtained as follows:
\begin{equation}
    \label{equ:batch_wise_statistics}
    \begin{aligned}
        \mu^{L}_{l} &= \frac{1}{BH_{l}W_{l}}\sum_{m=1}^{BH_{l}W_{l}} flat(z^L_{m,l}),\\
        (\sigma^L_{l})^2 &= \frac{1}{BH_{l}W_{l}}\sum_{m=1}^{BH_{l}W_{l}} (flat(z^L_{m,l}) - \mu^L_{l})^2,
    \end{aligned}
\end{equation}
where $flat(\cdot): \mathbb{R}^{B \times C_l \times H_l \times W_l} \rightarrow \mathbb{R}^{B H_l W_l \times C_l}$ indicates flattening operation, while $\mu_{l}^{L}, \sigma_{l}^{L} \in \mathbb{R}^{C_l}$ denote the mean and the variance of feature style, respectively. 

As thoroughly investigated in previous studies~\cite{li2016adabn, pan2018two}, these batch-wise statistics are highly related to the domain characteristics. In order to generate the domain statistics, we model the prior distributions of both style vectors~($\mu_{l}^{L}$, $\sigma_{l}^{L}$) as gaussians. For this purpose, we calculate the channel-wise mean and variance of style vectors as follows.
\begin{equation}
    \label{equ:channelwise_statistics}
    \begin{split}
        \hat{\mu}^L_{l} &= \frac{1}{C_{l}}\sum_{c=1}^{C_{l}} \mu_{c,l}^L,\;\;\;\;
        (\hat{\sigma}^L_{c,l})^2 = \frac{1}{C_{l}}\sum_{c=1}^{C_{l}} (\mu_{c,l}^L - \hat{\mu}^L_{l})^2,\\
        \tilde{\mu}^L_{l} &= \frac{1}{C_{l}}\sum_{c=1}^{C_{l}} \sigma_{c,l}^L,\;\;\;\;
        (\tilde{\sigma}^L_{c,l})^2 = \frac{1}{C_{l}}\sum_{c=1}^{C_{l}} (\sigma_{c,l}^L - \tilde{\mu}^L_{l})^2,
    \end{split}
\end{equation}
where $\hat{\mu}^L_{l}$ and $(\hat{\sigma}^L_{l})^2$ denote channel-wise statistics of $\mu_l^L$, while $\tilde{\mu}^L_{l}$ and $(\tilde{\sigma}^L_{l})^2$ are statistics of $\sigma_l^L$. $C_{l}$ is the number of channels of $z_{l}$.

To generate the novel domain styles, we manipulate the variance of distributions with the scaling parameters $s_\mu$ and $s_\sigma$, then sample new style vectors from its distribution as:
\begin{equation}
    \label{equ:sample_new_style}
    \begin{split}
        \mu^{\text{new}}_l \sim \mathcal{N}(\hat{\mu}^L_{l}, s_\mu (\hat{\sigma}^L_{l})^2), \\
        \sigma^{\text{new}}_l \sim \mathcal{N}(\tilde{\mu}^L_{l}, s_\sigma (\tilde{\sigma}^L_{l})^2).
    \end{split}
\end{equation}
As the variance increases, outlying style vectors, i.e., outliers, are more likely to be sampled from the distributions, whereas in-liers are sampled with higher probability in the opposite case. We show the effects of scale parameters through experiments in section~\ref{sec:analysis}.

After the sampling stage, the style vectors $\mu^{\text{new}}_l$ and $\sigma^{\text{new}}_l$ are applied to the original low frequency component $z^L_l$ via the affine transformation as follows.
\begin{equation}
    \label{equ:affine_transform}
        \Bar{z}^L_l = \sigma^{\text{new}}_l \left ( \frac{z^L_l - \mu^{L}_l}{\sigma^{L}_l} \right ) + \mu^{\text{new}}_l.
\end{equation}

We can interpret the sampling and transformation process as generating arbitrary domain statistics and applying on the original one. Notably, our affine transformation process is analogous to the batch normalization~(BN)~\cite{ioffe2015batch} where the affine parameters are $\mu^{\text{new}}_l$ and $\sigma^{\text{new}}_l$. Compared to adaptive instance normalization~(AdaIN)~\cite{huang2017arbitrary} which washes away discriminative features~\cite{nam2018batch, Seo2020LearningTO}, our feature stylization process conserves them while generating novel domain styles.

Lastly, the stylized low frequency feature $\Bar{z}^L_l$ are then combined with the original high frequency feature $z^H_l$ via the following equation.
\begin{equation}
    \label{equ:combine_components}
        \Bar{z}_l = z^H_l + \Bar{z}^L_l.
\end{equation}

We note that our feature stylization block can be inserted into any layer of the feature extractor $g$. Analyses on the best location $l$ of the feature stylization block will be discussed through ablation studies.

\begin{figure*}
  \includegraphics[width=1.0\textwidth]{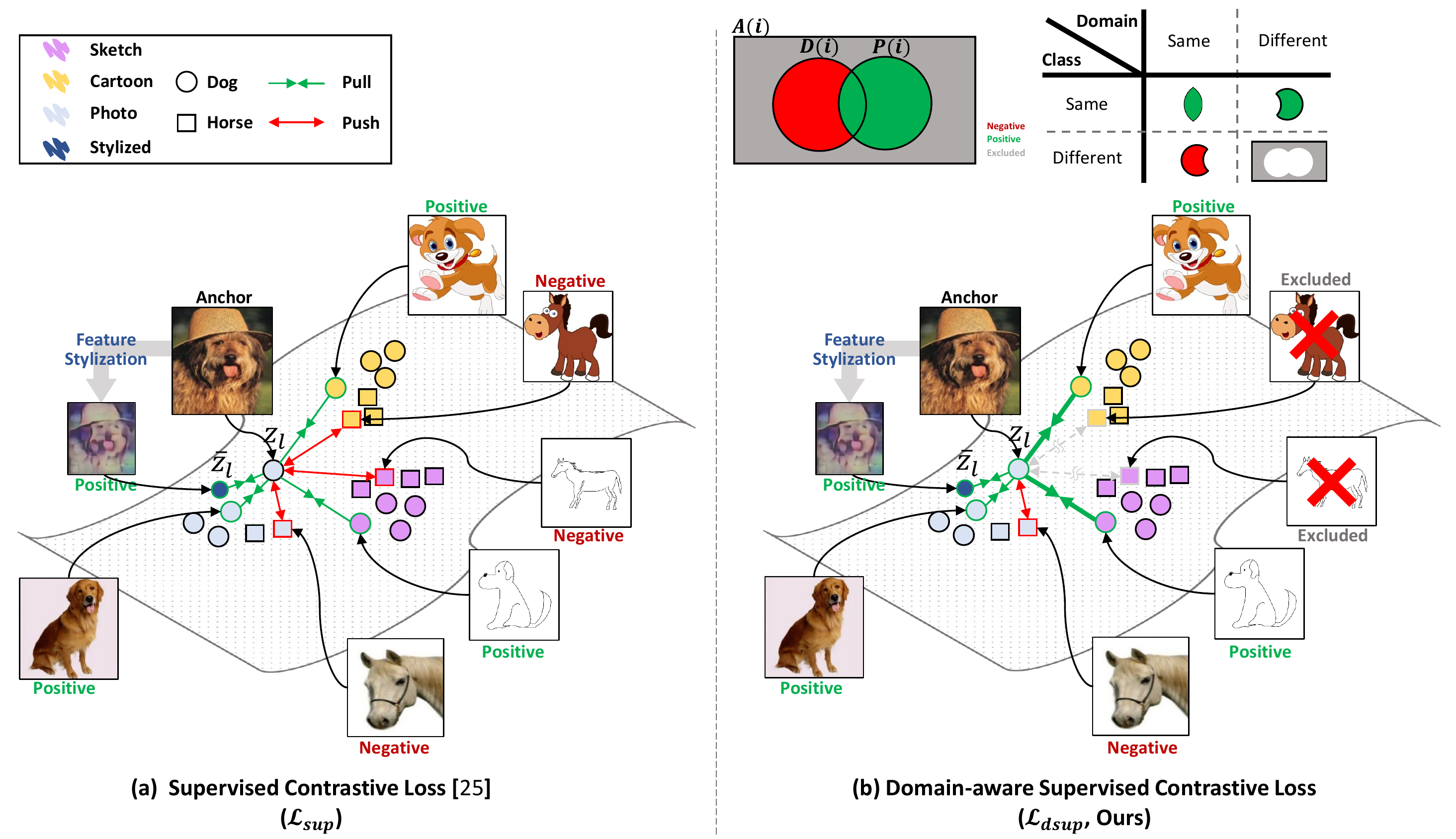}
  \caption{We illustrate the difference between the conventional supervised contrastive loss~\cite{khosla2020supervised}~($\mathcal{L}_{sup}$) and the proposed domain-aware supervised contrastive loss~($\mathcal{L}_{dsup}$). $\mathcal{L}_{sup}$ only considers the class label to compose the positive and negative sets. From the domain perspective, the sketch and cartoon domain images in $P(i)$ are attracted to the anchor, but those domains included in the negatives are expelled. This is contradictory since the positives contribute to the domain invariance while the negatives cause the domain discrepancy. For this, we propose $\mathcal{L}_{dsup}$ where we exclude the samples with different domains from the negative set. Consequently, the class discriminative feature is attainable and the domain-invariance is also accomplished by attracting positive samples from different domains.
  }
  \label{fig:dsup}
\end{figure*}

\subsection{Consistency Regularization}
The augmented feature $\Bar{z}_l$ is passed through the remaining layers, the same as the original feature $z_l$.
Given the stylized feature, we further encourage the model to output a consistent prediction with the original one.
For this, we minimize the discrepancy between output predictions from the original and the stylized features, which is formulated as:
\begin{equation}
    \label{equ:loss_consistency}
        \mathcal{L}_{cons} = - \frac{1}{K}\sum_{i=1}^{K} \frac{1}{n_i} \sum_{j=1}^{n_i}p(\Phi(x^i_j), \tau)\log(p(\Bar{\Phi}(x^i_j))), \\~0 \leq \tau \leq 1,
\end{equation}
where $p(\cdot)$ indicates the softmax function, $\Bar{\Phi}(x^i_j)$ denotes the neural network output in which feature stylization is performed on an intermediate layer, and $\tau$ is the temperature hyper-parameter.

The effect of the consistency loss is illustrated in~\Fref{fig:Arch}~(b). Through the consistency loss, the log-likelihood between the predictions is maximized. In addition, we apply temperature scaling with $\tau$ on the original prediction, denoted as $p(\Phi(x^i_j), \tau)$, to encourage the prediction of stylized feature to have low entropy~\cite{grandvalet2004semi, hinton2015distilling}. 

\subsection{Domain-aware Supervised Contrastive Loss}
Furthermore, we bring another intuition that a robust feature extractor should embed stylized features adjacent to original ones. Hence, we minimize the distance between the original feature (anchor) and the stylized one in terms of the dot-product similarity. This is accomplished by contrasting stylized features (positives), with other samples (negatives)~\cite{he2020momentum}. 

In addition, to encourage class discriminability, we adopt a supervised contrastive learning framework~\cite{khosla2020supervised} where output features from augmented samples and those from samples with the same class label are treated as positive. Meanwhile, the others in the mini-batch are considered negative.
The basic formulation of supervised contrastive learning is defined as:
\begin{equation}
    \label{equ:loss_sup}
        \mathcal{L}_{sup} = - \sum_{i \in I} \frac{1}{|P(i)|} \sum_{p \in P(i)}\log~\frac{exp(f'_i \cdot f'_p / \tau)}{\sum_{a \in A(i)}exp(f'_i \cdot f'_a / \tau)},
\end{equation}
where $I \equiv \{1 ... 2B\}$ indicates a set of indices of features and its stylized augmentation, $A(i) \equiv \left \{ I \setminus i \right \}$ contains the indices of all samples but the anchor, $P(i)$ denotes the set of indices of all positives to the anchor, and $f'$ denotes an output feature from the feature extractor $g$ after L2 normalization. The softmax function with temperature scaling is applied on the similarity matrix of the anchor.

As shown in \Fref{fig:dsup}~(a), the loss induces the model to attract positive features while repulsing negative ones from the anchor.
However, the performance degradation occurs when the loss is directly adopted for the the domain generalization task. Concretely, the feature space becomes domain-discriminative since the samples from different domains are pushed aside from the anchor. This is widely known to be detrimental for achieving the domain-invariance~\cite{daume2009frustratingly, sun2016return, tzeng2017adversarial, li2018domain}.
To this end, we propose to modify Eq.~\eqref{equ:loss_sup} to be more suitable for domain generalization, namely a \textit{domain-aware supervised contrastive loss}:

\begin{equation}
    \label{equ:loss_dsup}
            \mathcal{L}_{dsup} = - \sum_{i \in I} \frac{1}{|P(i)|} \sum_{p \in P(i)}\log~\frac{exp(f'_i \cdot f'_p / \tau)}{\sum_{a \in P(i) \cup D(i)}exp(f'_i \cdot f'_a / \tau)},
\end{equation}
where $D(i)$ is a set containing the indices of samples sharing the same domain label with the anchor.

As shown in \Fref{fig:dsup}~(b), $D(i)^c \cap P(i)^c$, \ie, samples from the different domain which were included in the earlier negative set, are excluded. For example, when a anchor $i$ is a ``photo dog'', its positive set $P(i)$ is $\{$``stylized photo dog'', ``cartoon dog'', ``sketch dog'', \dots $\}$, whereas remaining negatives belongs to different classes in ''photo'' domain.

Consequently, with the proposed domain-aware supervised contrastive loss, our feature extractor produces features not only discriminative to class labels but also invariant by attracting samples from different domain, i.e., $D(i)^c \cap P(i)$.

\subsection{Overall Training and Inference}

We train the neural network $\Phi$ with the weighted sum of losses as follows:
\begin{equation}
    \label{equ:loss_overall}
            \mathcal{L} = \mathcal{L}_{ce} + \lambda_{cons}\mathcal{L}_{cons} + \lambda_{dsup}\mathcal{L}_{dsup},
\end{equation}
where $\lambda_{*}$ is the weighting factor. The overall training is conducted in end-to-end manner. The network $\Phi$ is updated with respect to $\mathcal{L}_{ce}$ and $\mathcal{L}_{cons}$, while $\mathcal{L}_{dsup}$ affects only the feature extractor~$g$. During the inference, we detach the feature stylization module from the forward path so that the model predicts based on the original feature. 

\section{Experiments}
\label{sec:Experiments}

\subsection{Experiment Details}
\label{sec:experiment_details}
\textbf{Datasets.} As our evaluation benchmarks, we use PACS~\cite{Li2017DeeperBA} and Office-Home~\cite{finn2017model} following conventional settings. PACS is made up of four domains, \ie, Photo~(1,670 images), Art Painting~(2,048 images), Cartoon~(2,344 images) and Sketch~(3,929 images). The total number of images is 9,991 and the image resolution is $227 \times 227$. The dataset contains 7 common categories: `dog', `elephant', `giraffe', `guitar', `horse', `house', `person'.
Another benchmark is Office-Home which consists of images from four different domains, namely Artistic, Clip Art, Product, and Real world images. Each domain contains images of 65 object categories which are found in office and home. The total number of images is 15,500.

\noindent\textbf{Evaluation.} A common evaluation protocol in domain generalization is leave-one-domain-out evaluation~\cite{Li2017DeeperBA}. Specifically, we first select one domain as the target domain. Then, the other domains are set as source domains. We train our model on the source domains and evaluate it on the target domain. We note that any sample from the target domain is not allowed during the training step. This procedure is repeated to ensure that every domain is chosen to be the source domain exactly once, and we report the averaged accuracy.

\begin{table}[t]
  \caption{Quantitative leave-one-domain-out results on PACS. Entries are sorted in the chronological order and separated based on the backbones.}
  \label{tab:pacs}
  \centering
  \begin{tabular}{cccccc|c}
    \toprule
    &  & \multicolumn{5}{c}{Accuracy(\%)} \\
    & Method & Photo & Art & Cartoon & Sketch & Avg.\\
    \midrule
    \multirow{16}{*}{\rotatebox[origin=c]{90}{ResNet-18}}& Baseline~\cite{Li2017DeeperBA}& 95.19& 77.87& 75.89& 69.27 & 76.56 \\
    & D-SAM~\cite{d2018domain}& 95.30& 77.33& 72.43& 77.83 & 80.72\\
    & MetaReg~\cite{balaji2018metareg}& 95.50& 83.70& 77.20& 70.30 & 81.68\\
    & JiGen~\cite{carlucci2019domain}& 96.03& 79.42& 75.25& 71.35 & 80.51\\
    & MASF~\cite{dou2019domain}& 94.99& 80.29& 77.17& 71.69 & 81.04\\
    & Epi-FCR~\cite{Li2019EpisodicTF}& 93.90& 82.10& 77.00& 73.00& 81.50\\
    & InfoDrop~\cite{shi2020informative}& 96.11& 80.27& 76.54& 76.38 & 82.33\\
    & DMG~\cite{Chattopadhyay2020LearningTB}& 93.35& 76.90& 80.38& 75.21 & 81.46\\
    & EISNet~\cite{Wang2020LearningFE}& 95.93& 81.89& 76.44& 74.33 & 82.15\\
    & L2A-OT~\cite{zhou2020learning}& 96.20& 83.30& 78.20& 73.60& 82.83\\
    & DSON~\cite{Seo2020LearningTO}& 95.87& 84.67& 77.65& \textbf{82.23}& 85.11\\
    & RSC~\cite{Huang2020SelfChallengingIC}& 95.99& 83.43& 80.31& 80.85 & 85.15\\
    & MixStyle~\cite{zhou2021domain}& 96.10& 84.10& 78.80& 75.90 & 83.73\\
    & pAdaIN~\cite{nuriel2020permuted}& \textbf{96.29}& 81.74& 76.91& 75.13 & 82.52\\
    \cline{2-7}
    & \textit{Ours}& 95.63& \textbf{85.30}& \textbf{81.31}& 81.19&\textbf{85.86}\\
    \midrule
    \multirow{10}{*}{\rotatebox[origin=c]{90}{ResNet-50}}& Baseline~\cite{Li2017DeeperBA}& 97.66& 86.20& 78.70& 70.63 & 83.30 \\
    & MetaReg~\cite{balaji2018metareg}& 97.60& 87.20& 79.20& 70.30 & 83.58\\
    & MASF~\cite{dou2019domain}& 95.01& 82.89& 80.49& 72.29 & 82.67\\
    & DMG~\cite{Chattopadhyay2020LearningTB}& 94.49& 82.57& 78.11& 75.21 & 82.60\\
    & EISNet~\cite{Wang2020LearningFE}& 97.11& 86.64& 81.53& 78.07 & 85.84\\
    & DSON~\cite{Seo2020LearningTO}& 95.99& 87.04& 80.62& 82.90 & 86.64\\
    & RSC~\cite{Huang2020SelfChallengingIC}& \textbf{97.92}& 87.89& 82.16& \textbf{83.35}&87.83\\
    & pAdaIN~\cite{nuriel2020permuted}& 97.17& 85.82& 81.06& 77.37 & 85.36\\
    \cline{2-7}
    & \textit{Ours}& 96.59& \textbf{88.48}& \textbf{83.83}& 82.92&\textbf{87.96}\\
  \bottomrule
  \end{tabular}
\end{table}

\noindent\textbf{Implementation details.} For fair comparison with previous studies, we adopt ResNet-18 and ResNet-50~\cite{he2016deep} pre-trained on ImageNet~\cite{krizhevsky2012imagenet}. We optimize our network with Stochastic Gradient Descent (SGD) optimizer. We set an initial learning rate as 0.004 and train for 40 epochs. The decay rate is set to 0.0005 which is applied after 20 epochs. A single mini-batch contains a total of 126 images, 42 images for each source domain. 

Inspired by FixMatch~\cite{sohn2020fixmatch} and SupCon~\cite{khosla2020supervised}, the temperature parameters $\tau$ of $\mathcal{L}_{cons}$ and $\mathcal{L}_{dsup}$ are set to 0.5 and 0.15, respectively. Considering the scale of loss fucntions, weighting factors ($\lambda_{cons}$, $\lambda_{dsup}$) are set to (0.3, 12) and (0.9, 6) for ResNet-18 and ResNet-50, respectively. Besides, the scale parameters ($s_{\mu}, s_{\sigma}$) are set to 10 and 20 for ResNet-18 and ResNet-50, respectively. Our model is built upon the popular implementation of Zhou et al.~\cite{zhou2020domain}.


\subsection{Comparison with State-of-the art Methods}
\noindent\textbf{Results on PACS.} In \Tref{tab:pacs}, we compare our method with previous domain generalization methods on PACS dataset~\cite{Li2017DeeperBA}. Recognizably, our method beats previous approaches and achieves a new state-of-the-art performance with the average accuracy of 85.86\% with ResNet-18.
Consistently, our method shows large improvements when adopting ResNet-50 as our backbone, accomplishing a new record with the average accuracy of 87.86\% across the leave-one-domain-out scenarios. 

Through the experiments, vivid performance gains are observed when art and sketch domains are set as target domains. This is reasonable since our method has the strength in generating novel styles while preserving the shape cues which are essential for accurate classification in those domains.
Despite the slight performance degradation in the photo domain, our method outperforms the competitors in terms of the average accuracy, validating the better domain generalization ability.
We note that our feature stylization module does not require additional network parameters, which makes our model more competitive in terms of memory. 

\begin{table}[t]
  \caption{Quantitative leave-one-domain-out results on Office-home. Entries are sorted in the chronological order.}
  \label{tab:office}
  \centering
  \begin{tabular}{cccccc|c}
    \toprule
    &  & \multicolumn{5}{c}{Accuracy(\%)} \\
    & Method & Art & Clipart & Product & Real & Avg.\\
    \midrule
    \multirow{11}{*}{\rotatebox[origin=c]{90}{ResNet-18}}& Baseline~\cite{venkateswara2017deep}& 52.15& 45.86& 70.86& 73.15 & 60.51 \\
    & CCSA~\cite{Motiian2017UnifiedDS}& 59.90& 49.9& 74.10& 75.7 & 64.90\\
    & D-SAM~\cite{d2018domain}& 58.03& 44.37& 69.22& 71.45 & 60.77\\
    & CrossGrad~\cite{shankar2018generalizing}& 58.40& 49.40& 73.90& 75.80 & 64.38\\
    & MMD-AAE~\cite{li2018domain}& 56.50& 47.30& 72.10& 74.80 & 62.68\\
    & JiGen~\cite{carlucci2019domain}& 53.04& 47.51& 71.47& 72.79 & 61.20\\
    & L2A-OT~\cite{zhou2020learning}& \textbf{60.60}& 50.10& \textbf{74.80}& \textbf{77.00} & 65.63\\
    & DSON~\cite{Seo2020LearningTO}& 59.37& 45.70& 71.84& 74.68 & 62.90\\
    & RSC~\cite{Huang2020SelfChallengingIC}& 58.42& 47.90& 71.63& 74.54 & 63.12\\
    & MixStyle~\cite{zhou2021domain}& 58.70& 53.40& 74.20& 75.90 & 65.55\\
    \cline{2-7}
    & \textit{Ours}& 60.24& \textbf{53.54}& 74.36& 76.66&\textbf{66.20}\\
  \bottomrule
\end{tabular}
\end{table}

\begin{table}
  \caption{Results of single source domain generalization on PACS. Each row and column indicates the source and target domain, respectively. We report the accuracy with the absolute gain from baseline in brackets. Positive and negative gains are colored \textcolor{darkpastelgreen}{green} and \textcolor{red}{red}, respectively.}
  \label{tab:single domain generalization}
  \centering
\resizebox{0.8\linewidth}{!}
{
    \begin{tabular}{c|cccc}
    \toprule
    & \multicolumn{4}{c}{\begin{tabular}[c]{@{}c@{}}Accuracy~(\%)\\ (Absolute gain from baseline)\end{tabular}} \\
    & Photo & Art painting & Cartoon & Sketch\\
    \midrule
    Photo & \begin{tabular}[c]{@{}c@{}}99.88\\ (+0.00)\end{tabular} & \begin{tabular}[c]{@{}c@{}}63.18\\ (+\textcolor{darkpastelgreen}{5.61})\end{tabular} & \begin{tabular}[c]{@{}c@{}}21.84\\ (+\textcolor{darkpastelgreen}{2.43})\end{tabular} & \begin{tabular}[c]{@{}c@{}}54.71\\ (+\textcolor{darkpastelgreen}{28.59})\end{tabular}\\
    Art painting& \begin{tabular}[c]{@{}c@{}}96.53\\ (-\textcolor{red}{0.06})\end{tabular} & \begin{tabular}[c]{@{}c@{}}99.46\\ (+0.00)\end{tabular} & \begin{tabular}[c]{@{}c@{}}67.79\\ (+\textcolor{darkpastelgreen}{11.56})\end{tabular} & \begin{tabular}[c]{@{}c@{}}53.59\\ (+\textcolor{darkpastelgreen}{9.60})\end{tabular} \\
    Cartoon & \begin{tabular}[c]{@{}c@{}}84.97\\ (+\textcolor{darkpastelgreen}{0.54})\end{tabular} & \begin{tabular}[c]{@{}c@{}}70.56\\ (+\textcolor{darkpastelgreen}{9.09})\end{tabular} & \begin{tabular}[c]{@{}c@{}}99.57\\ (+\textcolor{darkpastelgreen}{0.12})\end{tabular} & \begin{tabular}[c]{@{}c@{}}70.90\\ (+\textcolor{darkpastelgreen}{8.50})\end{tabular} \\
    Sketch & \begin{tabular}[c]{@{}c@{}}41.56\\ (+\textcolor{darkpastelgreen}{9.16})\end{tabular} & \begin{tabular}[c]{@{}c@{}}43.07\\ (+\textcolor{darkpastelgreen}{13.53})\end{tabular} & \begin{tabular}[c]{@{}c@{}}60.20\\ (+\textcolor{darkpastelgreen}{15.96})\end{tabular} & \begin{tabular}[c]{@{}c@{}}99.36\\ (-\textcolor{red}{0.08})\end{tabular} \\
  \bottomrule
\end{tabular}
}
\end{table}

\noindent\textbf{Results on office-Home.} We also provide the results on Office-Home benchmark~\cite{finn2017model} in \Tref{tab:office}. Again, our method breaks the record with ResNet-18, achieving the average accuracy of 66.2\%. Conspicuously, ours makes the performance improvements regardless of the target domain. Overall comparisons verify the effectiveness of our feature stylization strategy and the proposed contrastive loss.

\subsection{Single-source Domain Generalization}

In \Tref{tab:single domain generalization}, we present the single source domain generalization results with ResNet18 on PACS benchmark.
In this setting, only a single domain is selected as a source dataset and the trained model is tested on other target domains.
Rows and columns indicate source and target domains, respectively.
We use the same hyper-parameter settings described in the previous section, except for the scale parameters $s_{\mu}$, $s_{\sigma}$ both of which are scaled down to 5. In addition, since the source domain is single in this setting, $D(i)$ in $\mathcal{L}_{dsup}$ is naturally ignored.

Except for the diagonal elements where train and test domains are the same, we achieve improvements on the most of domain generalization scenarios. We can observe remarkable performance gain on ``Photo-to-Sketch'', ``Art-to-Cartoon'', ``Sketch-to-Art'', and ``Sketch-to-Cartoon'' settings. Especially, a huge performance improvement is observed when the sketch is used for the source domain, \ie, only coarse shape information is available for training. This demonstrates the style diversification ability of our feature stylization block.
The ``Art-to-Photo'' setting is the only generalization scenario where a performance degradation is observed but still in an acceptable margin.

\subsection{Analysis}
\label{sec:analysis}
In this section, we analyze our method and conduct ablation studies on the PACS benchmark with ResNet-18 backbone. To be specific, we first analyze the effect of each loss function, then we investigate the correlation between the performance and the scale parameter $s$. Thereafter, we examine the most suitable location of the feature stylization block followed by the analysis on the feature decomposition with frequency components. We note that all remaining hyper-parameters are fixed through ablation studies.

\textbf{Ablation study on components.}
We conduct an ablation study to inspect the contribution of the feature stylization along with loss functions. As shown in Table~\ref{tab:ablation loss}, every single component enlarges the generalization capacity of the model compared to the baseline. In detail, the effectiveness of feature stylization is observable with the overall performance improvement of $\sim$4.08\%, in terms of the average accuracy. Specifically, the performance gain on the sketch domain is delightful, reaching $\sim$8.71\%. In addition, it is verified that the consistency loss boosts domain robustness by regularizing the output discrepancy between original and stylized features. Moreover, the proposed domain-aware contrastive loss enhances the performance by pursuing feature similarities between different domains but with the same category. Consequently, with the harmonious combination of aforementioned components, we can find that all components are complementary and have a positive effect in the domain generalization task.

\begin{table}[t]
    \caption{Ablation study on each component. $\dagger$ denotes that stylized features are aggregated for cross-entropy loss~(${\mathcal{L}_{ce}}$).}
    \label{tab:ablation loss}
    \centering
    \resizebox{0.91\linewidth}{!}{
      \begin{tabular}{ccc|cccc|c}
        \toprule
        \multicolumn{3}{c}{} & \multicolumn{5}{|c}{Accuracy(\%)} \\
        \multicolumn{1}{c}{\begin{tabular}[c]{@{}c@{}}\small{Feature}\\ \small{Transform}\end{tabular}} & $\mathcal{L}_{cons}$ &$\mathcal{L}_{dsup}$& P & A & C & S & Avg.\\
        \midrule
        & & & \textbf{96.29} & 76.90 & 77.30 & 68.81 & 79.83\\
        ~\checkmark$^{\dagger}$& & & 96.11 & 82.03 & 80.12 & 77.52 & 83.95\\
        \checkmark& \checkmark& & 95.45 & 84.18 & 80.08 & 79.30 & 84.75\\
        \checkmark& & \checkmark& 95.15 & 80.42 & 80.59 & 73.75 & 82.48\\
        \checkmark& \checkmark& \checkmark& 95.63 & \textbf{85.30} & \textbf{81.31} & \textbf{81.19} & \textbf{85.86}\\
      \bottomrule
    \end{tabular}
}
\end{table}

\begin{figure}
  \includegraphics[width=0.48\textwidth]{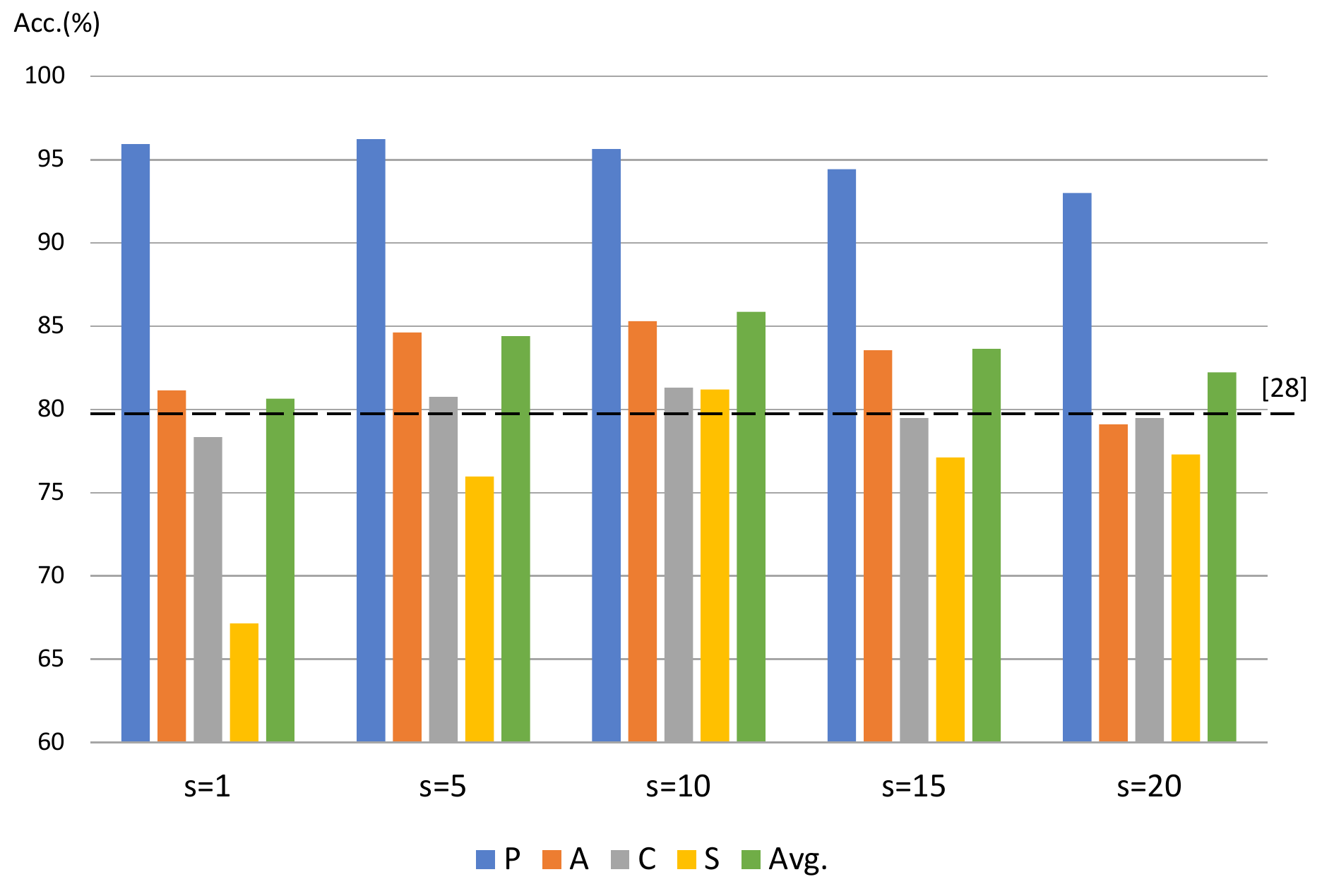}
  \caption{Ablation of scale parameter. The value of scale parameters and the accuracy is on the x and y axis, respectively. We also draw the average accuracy of baseline as a dotted line for better comparison.}
  \label{fig:abl scale parameter}
\end{figure}

\textbf{The scale parameter.}
In \Fref{fig:abl scale parameter}, we compare different scale parameters in the feature stylization block. We adjust scale parameters~($s_{\mu}$, $s_{\sigma}$) in $\left \{ 1, 5, 10, 15, 20\right \}$. With the scale parameters of 1, augmented style vectors follow the original style distribution, leading to a marginal improvement. As the scale parameter increases, the outlying style vectors are more likely to be sampled, thus increasing the generalization ability. However, excessive scale values produce too distant features and outputs from the original one, resulting in high favoritism on shape cues. This is undesirable since shape cues are not sufficient for visual recognition and texture cues still contains class-discriminative information as in the human visual system~\cite{sann2007perception}. The best performance is found at the scale parameter of 10, which may be the ``sweet spot'' of exploiting both shape and textural cues.

\textbf{Feature decomposition strategy.}
We verify the effect of decomposing the feature into high frequency and low frequency components.
In \Tref{tab:ablation fd}, we compare between exploiting whole feature without decomposition (-), high frequency components $z^H$, and low frequency components $z^L$ for feature stylization.
As shown in \Tref{tab:ablation fd}, the best performance is achieved when the feature stylization is applied only on low frequency components. Applying stylization on high frequency feature falls behind the other strategies, since only the shape information is partially distorted. Meanwhile, although transforming the whole feature seems to be a good strategy overall, it shows inferior performances on the domains where shape cues are crucial, such as art and sketch domains.

\begin{table}
  \caption{Ablation study on feature decomposition strategies. The column ``Frequency component'' denotes the component where feature stylization is applied. ``-'' denotes use of the original feature $z$, while $z^{H}$ and $z^{L}$ indicate the high frequency and low frequency features respectively.}
  \label{tab:ablation fd}
  \centering
\resizebox{0.97\linewidth}{!}
{
  \begin{tabular}{cc|cccc|c}
    \toprule
    &  &\multicolumn{5}{c}{Accuracy(\%)} \\
    \multicolumn{1}{c}{\begin{tabular}[c]{@{}c@{}}Frequency\\ component\end{tabular}} &$\mathcal{L}_{cons}$, $\mathcal{L}_{dsup}$& P & A & C & S & Avg.\\
    \midrule
    -& & 95.45 & 81.79 & 79.44 & 77.47 & 83.54\\
    $z^{H}$& & 96.05 & 77.34 & 78.58 & 72.73 & 81.18\\
    $z^{L}$~(\textit{Ours})& & 96.11 & 82.03 & 80.12 & 77.52 & 83.95\\
    \midrule
    -& \checkmark& 95.87 & 83.98 & 81.06 & 80.30 & 85.30\\
    $z^{H}$& \checkmark& \textbf{96.41} & 78.61 & 79.78 & 72.25 & 81.76\\
    $z^{L}$~(\textit{Ours})& \checkmark& 95.63 & \textbf{85.30} & \textbf{81.31} & \textbf{81.19} & \textbf{85.86}\\
  \bottomrule
\end{tabular}
}
\end{table}

\textbf{Location of the feature stylization block.}
We discuss on where the proposed feature stylization block should be located.
We denote these stack of residual blocks by re-grouping the ResNet architecture into 5 groups of layers, {\fontfamily{qcr}\selectfont Conv} and {\fontfamily{qcr}\selectfont ResBlock 1-4}, where {\fontfamily{qcr}\selectfont Conv} denotes the first convolutional layer before residual blocks. 

As shown in Table~\ref{tab:ablation ft}, the best spot of the proposed feature stylization is right after the second residual blocks. This is quite reasonable considering the nature of deep neural networks~\cite{gatys2016image, donahue2014decaf, shi2020informative}, where features at this level adequately represent low-level structural information as well as high-level semantic information.

\begin{table}
  \caption{Ablation study on the location of feature stylization block.}
  \label{tab:ablation ft}
  \centering
\resizebox{0.86\linewidth}{!}
{
\begin{tabular}{c|cccc|c}
    \toprule
    & \multicolumn{5}{c}{Accuracy(\%)} \\
    Layer & P & A & C & S & Avg.\\
    \midrule
    \fontfamily{qcr}\selectfont Conv & 94.91 & 81.25 & 79.10 & 77.85 & 83.28\\
    \fontfamily{qcr}\selectfont ResBlock${_{1}}$ & \textbf{96.11} & 83.74 & 80.33 & 76.02 & 84.05\\
    \fontfamily{qcr}\selectfont ResBlock${_{2}}$ & 95.63 & \textbf{85.30} & \textbf{81.31} & \textbf{81.19} & \textbf{85.86}\\
    \fontfamily{qcr}\selectfont ResBlock${_{3}}$ & 95.99 & 82.19 & 78.67 & 68.13 & 81.24\\
    \fontfamily{qcr}\selectfont ResBlock${_{4}}$ & 95.39 & 82.18 & 79.14 & 70.77 & 81.87\\
  \bottomrule
\end{tabular}%
}
\end{table}

\section{Conclusion}
In this paper, we proposed a novel framework for domain generalization, where the features are stylized into diverse domains. In detail, we sampled domain style vectors from the manipulated distribution of batch-wise feature statistics, then utilized the style vectors for affine transformation.
To achieve the domain robustness, we exploited stylized features for regularization in terms of output consistency and feature similarity via consistency loss and novel domain-aware supervised contrastive loss, respectively. 
Through comparisons and extensive analyses on two popular benchmarks, we demonstrated the effectiveness of the proposed feature stylization and two losses.

\begin{acks}
This research was partly supported by the MSIT (Ministry of Science, ICT), Korea, under the High-Potential Individuals Global Training Program (No. 2021-0-01696) supervised by the IITP (Institute for Information \& Communications Technology Planning \& Evaluation), the National Research Foundation of Korea (NRF) grant funded by the Korea government (MSIT) (No. 2019R1A2C2003760), and the Institute for Information \& Communications Technology Planning \& Evaluation~(IITP) grant funded by the Korea government (MSIT) (No. 2020-0-01361: Artificial Intelligence Graduate School Program (YONSEI UNIVERSITY)). This project was also supported by Microsoft Research Asia.
\end{acks}

\bibliographystyle{ACM-Reference-Format}
\bibliography{acmart.bib}










\end{document}